\documentclass[runningheads]{llncs}
\usepackage{graphicx}
\usepackage{url}
\usepackage{color}
\usepackage{multirow}
\usepackage{booktabs}
\usepackage{array}
\usepackage{mathrsfs}
\usepackage{subfig}

\usepackage{mathtools}
\usepackage{amssymb}
\usepackage{fancyhdr}
\usepackage{CJKutf8}

\begin{document}
\titlerunning{Joint Event Causality Extraction via Convolutional Semantic Infusion}
\title{Back to Prior Knowledge: Joint Event Causality Extraction via Convolutional Semantic Infusion}

\author{Zijian Wang\and
Hao Wang\inst{*} \and
Xiangfeng Luo\inst{*} \and
Jianqi Gao}
\institute{School of Computer Engineering and Science, Shanghai University, Shanghai 200444, China \\
\email{\{zijianwang,wang-hao,luoxf,gjqss\}@shu.edu.cn}\\
}
%
%\authorrunning{Wang et al.}
\authorrunning{Zijian Wang et al.}
% \institute{XXX University, Princeton NJ 08544, USA \and
\maketitle              % typeset the header of the contribution
\begin{abstract}
Joint event and causality extraction is a challenging yet essential task in information retrieval and data mining. Recently, pre-trained language models (e.g., BERT) yield state-of-the-art results and dominate in a variety of NLP tasks. However, these models are incapable of imposing external knowledge in domain-specific extraction. Considering the prior knowledge of frequent n-grams that represent cause/effect events may benefit both event and causality extraction, in this paper, we propose convolutional knowledge infusion for frequent n-grams with different windows of length within a joint extraction framework. Knowledge infusion during convolutional filter initialization not only helps the model capture both intra-event (i.e., features in an event cluster) and inter-event (i.e., associations across event clusters) features but also boosts training convergence. Experimental results on the benchmark datasets show that our model significantly outperforms the strong BERT+CSNN baseline.

\keywords{Causality extraction \and Prior knowledge \and Semantic infusion.}
\end{abstract}
\section{Introduction}
Joint event and causality extraction from natural language text is a challenging task in knowledge discovery  \cite{asghar2016automatic}, discourse understanding \cite{li2019knowledge}, and machine comprehension \cite{sorgente2013automatic}. Formally, joint event causality extraction is defined as a procedure of extracting an event triplet consisting of a cause event $e1$, an effect event $e2$ with the underlying causality relationship. Due to the complexity and ambiguity of natural languages, event causality extraction remains a hard problem in case no high-quality training dataset is not available like the financial domain.

Conventional methods  commonly treat event causality extraction as a two-phase process in a pipeline manner, divided into event mining and relation inference \cite{zhao2016event}. However, the pipeline approach suffers from error propagation. Incorporating the current neural model with the prior knowledge in existing domain-specific knowledge bases is complicated.

Recently, neural-based extraction methods have become a majority of related tasks in NLP, relation classification \cite{shen2016attention}, relation extraction \cite{lin2016neural} and sequence tagging \cite{jin2020inter}. Among these methods, pre-trained language models \cite{devlin2018bert} (e.g., BERT) dominate the state-of-the-art results on a wide range of NLP tasks. BERT provides a masked language model (MLM) supporting fine-tuning to achieve better performance on a new dataset while reducing serious feature engineering efforts. However, there are still drawbacks when applying such pre-trained language models for event causality extraction as follows:
\begin{itemize}
\item The small size of available datasets is the main bottleneck of fine-tuning BERT-based models to satisfactory performance.
\item Even if BERT trained on a large-scale corpus contains commonsense knowledge, it may not be sufficient for specific-domain such as financial.
\item Domain-specific frequency analysis of n-grams should be essential prior knowledge to recognize events. In contrast, these important hints have not been fully emphasized in the current neural architecture.
\end{itemize}
%Bert in the Chinese domain is based on character granularity, and a large amount of semantic information in Chinese is contained in words. Models based solely on word granularity tend to lose semantic information. On the other hand, ，the Bert model has high requirements for hardware.

% Initialisation techniques play an essential role of catalyst for the revival of neural networks. Considering the embedding layers could be initialised by pre-trained word vectors, intuitively, weights in other layers that are still randomly initialised?

\begin{figure}[t]
\centering
\includegraphics[width=1.0\linewidth]{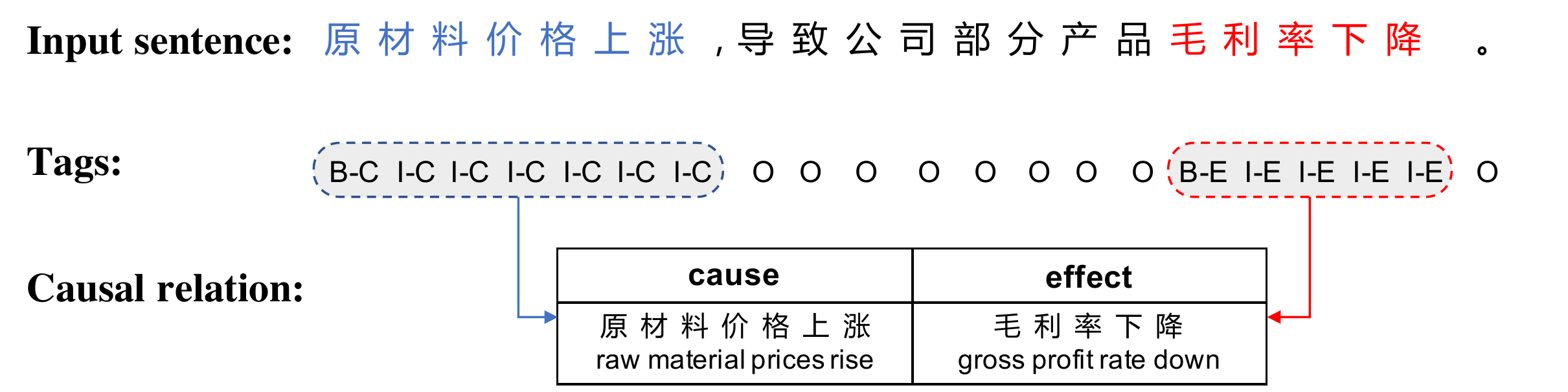}
\caption{An example of joint event causality extraction from raw financial text. It extracts cause and effect events with the underlying cause-effect relation simultaneously.}
\label{fig00}
\end{figure}

To tackle these issues, in this paper, we propose a novel joint method for event causality extraction. We first pass the input text through the BERT encoding layer to generate token representations. Secondly, to infuse the knowledge like frequent cause/effect n-grams on different scales, we utilize multiple convolutional filters simultaneously (i.e., infusing intra-event knowledge). The weights of these filters are manually initialized with the centroid vectors of the cause/effect n-gram clusters. After that, we link the cause and effect events using the key-query attention mechanism to alleviate incorrect cause-effect pair candidates (i.e., infusing inter-event knowledge). Finally, we predict target labels given the contextual representation by combing bidirectional long short-term memory (LSTM) with the conditional random field (CRF). Empirical results show that our model fusing advantages of both intra- and inter-event knowledge significantly improve event causality extraction and obtain state-of-the-art performance.

The contributions of this paper can be summarized as follows:
\begin{enumerate}
\item We propose a novel joint framework of event causality extraction based on recent advances of the pre-trained language models, taking account of frequent domain-specific and event-relevant n-grams given statistic analysis.
\item This framework allows incorporating intra-n-gram knowledge of frequent n-grams into the current deep neural architecture to filter potential cause or effect event mentions in the text during extracting.
\item Our approach also considers inter-n-gram knowledge of cause-effect co-occurrence. We adopt a query-key attention mechanism to pairwisely extract cause-effect pairs.
\end{enumerate}
\section{Related Work}
The methods for event or causality extraction in the literature fall into three categories: rule-based, machine learning, and neural network. The rule-based methods employ linguistic resources or NLP toolkits to perform pattern matching. These methods often have low adaptability across domains and require extensive in-domain knowledge to deal with the domain generalisation problem. The second category methods are mainly based on machine learning techniques, requiring considerable human effort and additional time cost in feature engineering. These methods heavily rely on manual selection of textual feature sets. The third category methods are depending on the neural network. In this section, we survey those methods and figure out the problems existing in those methods.

\subsection{Pattern Matching}
Numerous rule-based methods have been dedicated to event causality mining. Early works predominantly perform syntactic pattern matching, where the patterns or templates are handcrafted for the specific-domain texts. For instance, Grishman et al. \cite{grishman1988domain} perform syntactic and semantic analysis to extract the temporal and causal relation networks from texts. Kontos et al. \cite{kontos1991acquisition} match expository text with structural patterns to detect causal events. Girju et al. \cite{girju2002text} validate acquired patterns in a semi-supervised way. They check whether to express a causal relationship based on constraints of nouns and verbs. As a result, those methods cannot generalize to a variety of domains.

\subsection{Machine Learning}
The paradigm of automatic causal extraction dates back to machine learning techniques using trigger words combined with decision trees \cite{riaz2010another} to extract causal relations. Sorgente et al. \cite{sorgente2013automatic} first extract the candidates of causal event pairs with pre-defined templates and then use a Bayesian classifier to filter non-causal pairs. Zhao et al. \cite{zhao2016event} compute the similarity of syntactic dependency structures to integrate causal connectives. Indeed, these methods suffer from data sparsity and require professional annotators.

\subsection{Neural Network}
Due to the powerful deep neural representations, neural networks can effectively extract implicit causal relations. In recent years, the adoption of deep learning techniques for causality extraction has become a popular choice for researchers. Methods of event extraction can be roughly divided into the template argument filling approach or sequence labelling approach.
For example, Chen et al. \cite{yubo2015event} propose to process extraction as a serial execution of relation extraction and event extraction. This method splits the task into two phases. They first extract the event elements, then put them into a pre-defined template consisting of event elements. By sharing features during extraction, it effectively avoids missing the vital information.

Other models obey a sequential labelling manner. Fu et al. \cite{jian2011event} propose to treat the causality extraction problem as a sequential labelling problem. Martinez et al. \cite{martinez2017neural} employ an LSTM model to make contextual reasoning for predicting event relation. Jin et al. \cite{jin2020inter} propose a cascaded network model capturing local phrasal features as well as non-local dependencies to help cause and effect extraction. Despite their success, those methods ignore the fact of domain-specific extraction. On the one hand, an n-gram representing cause or effect event should frequently appear in the text. On the other hand, the events involving causality have a higher probability of co-occurrence.

\section{Our Model}
%In sequence tagging tasks, the unit is usually a single character or word, while the n-gram-level connection is ignored. From the perspective of statistics, n-grams describing events commonly have a high frequency and often co-occur in the sentence. Therefore, in our proposal, we incorporate the prior information of causal events from the dataset into the model. The knowledge of n-gram length is used as the guidance of event granularity.

\begin{figure}[t]
\centering
\includegraphics[width=1.0\linewidth]{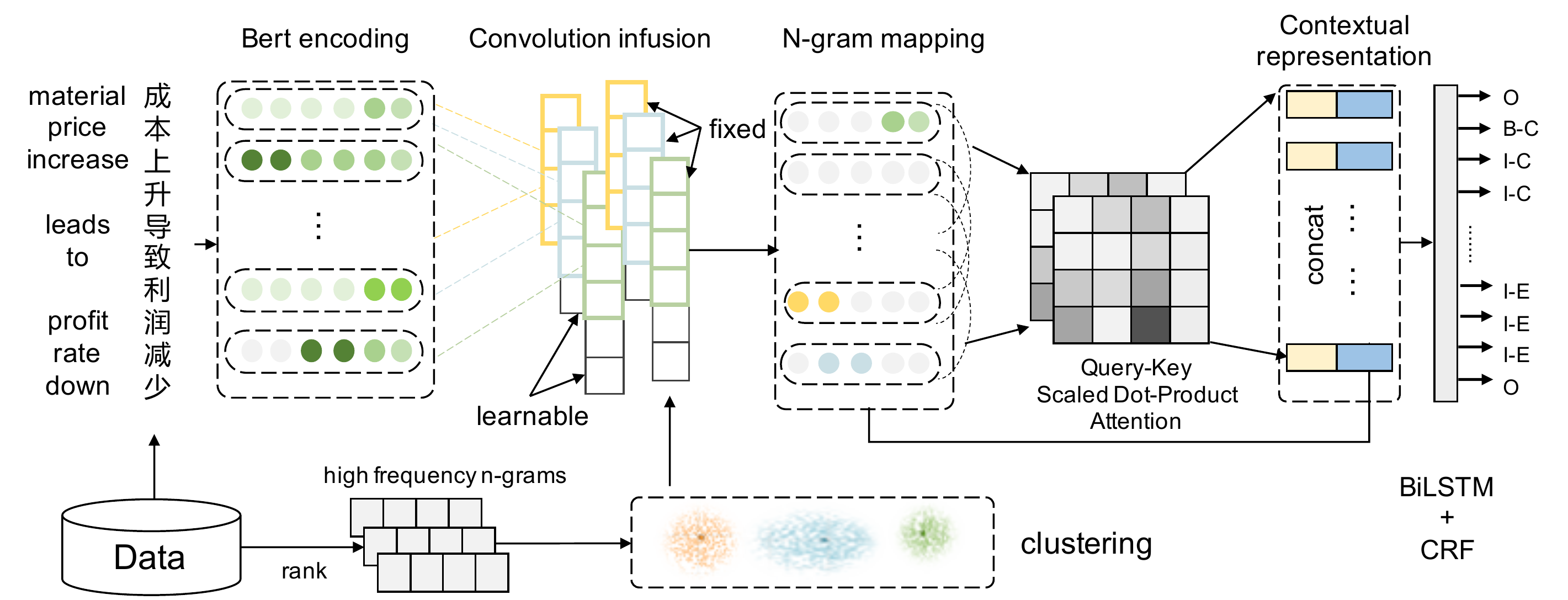}
\caption{Illustration of the model architecture. Intra- and inter-event feature are extracted using convolutional knowledge infusion and key-query attention mechanism. This model allows to extract cause and effect events associated with a event causal relation at the same time. }
\label{fig01}
\end{figure}

Figure~\ref{fig01} shows the overall architecture of the model. It is composed of four modules: 1) We first pass the input sentence through the BERT encoding layer and the alignment layer successively to output the BERT representation for each token. 2) A convolutional knowledge infusion layer is created to capture useful n-gram patterns to focus on domain-relevant n-grams at the beginning of the training phase. 3) We characterize inter-associations among cause and effect using a key-query attention mechanism. 4) Temporal dependencies are established utilising the LSTM combined with CRF layers, which significantly improves the performance of causality extraction.

\subsection{BERT Encoder}
The BERT is one of the pre-trained language models, which contains multi-layers of the bidirectional transformer, designed to jointly condition on both left and right context. BERT supports fine-tuning for a wide range of tasks without substantial task-specific architecture modifications. We omit the exhaustive background description of the architecture of BERT in this paper.

In our model, we use the BERT encoder to model sentences that contain the event mentions via computing a context-aware representation for each token We take the packed sentences $[CLS,S,SEP]$ as the input. ``[CLS]'' is inserted as the first token of each input sequence. ``[SEP]'' denotes the end of the sentence. For each token $s_i$ in $S$, the BERT input can be represented as:
\begin{equation}
s_i= [s_i^{tok}\oplus s_i^{pos} \oplus s_i^{seg}],
\end{equation}
where $s^{tok}_i$, $s^{pos}_i$, and $s^{seg}_i$ represent token, position, and segment embeddings for $i$, respectively. In this regard, a sentence is expressed as a vector $H \in \mathbb{R}^{l\times e}$, where $l$ is the length of sentence and $e$ is the size of embedding dimension. $h_i$ is a vector standing for word embedding of the $i$-th word in the sentence.
\begin{equation}
h^i_N=BERT([CLS,s_1,\dots, s_i,\dots,s_l, SEP]),
\end{equation}
where $h^i_N$ represents hidden states of the last layer generated by BERT. We use them as the context-aware representation for each token.

\subsection{Convolutional Semantic Infusion}
In natural language processing, the convolutional operation can be understood as a process in which a sliding window continuously captures semantic features in a sentence of specific length. The convolution operation will extract semantic features similar to the convolution filter (i.e., the vectors similar to the convolution filter should share a higher weight).
In our model, the convolution layer aims to enhance event-relevant features by embedding causal patterns into feature maps. Considering that word embedding can be initialized using pre-trained word vectors, it also should be possible to initialize the convolutional filter (i.e., kernel) with the vector related to frequent event n-grams. Therefore, inspired by \cite{li2017initializing}, we launch unsupervised clustering to collect n-grams mentioning the similar cause or effect events in the text into the same clusters. Thus, these clusters can elegantly represent cause/effect semantic. Then, we compute the centroid vector of the cluster given all cause/effect event n-grams in the cluster and initialize convolution filters with these vectors. We only fix a part of weight in the filter using the cluster semantic vector, which allows our model to learn more features by itself during training. The details are given below:

\subsubsection{Ngram Collection:} Sentences can be segmented into particular chunks according to the size of the window in the convolution operation, e.g.,
\begin{CJK*}{UTF8}{gbsn}
``我喜欢苹果(I like apples)'' are split into four bigrams ``我喜,喜欢,欢苹,苹果'' when n=2, step=1).
\end{CJK*}
This method takes advantage over standard word representations to obtain ampler semantic information, which motivates us to recognize the event phrase boundary in this example.

The texts in a different domain often have different length for event n-grams, which also dominates convolutional window size. At the same time, the n-gram length is a kind of vital prior knowledge in event extraction. However, it is often ignored by conventional character- or word-based methods, resulting in errors of event boundary recognition. Therefore, we count the event length in the dataset. We find the number of the events in the most frequent length accounted for 27\% among all examples in the Financial dataset. Motivated by this phenomenon, we associate the n-gram length feature as well as n-gram clusters by applying the convolution filters with various window sizes.

Intuitively, ``profit drop'' should be more important than the location n-gram like ``New York'' in most cases in the financial domain. Thus, we distill the relevant n-grams in a cluster with the centroid vector. In practical, we collect n-grams only from the training data. For its simplicity and effectiveness, we sort the n-grams using Na\"ive Bayes to obtain scores. The ranking score $r$ of the n-gram $w$ is calculated as follows:
\begin{equation}
r = \frac{\left(p_c^w+b \right) / {\left \| \ p_c \right \|}_1}{\left(p_e^w+b \right) / {\left \| \ p_e \right \|}_1},
\label{e3}
\end{equation}
where $c$ denotes cause event and $e$ indicates effect event. $p_c^w$ is the number of sentences that contains n-grams $w$ in class $c$. ${\left \| p_c \right \|}_1$ is the number of n-grams related to cause $c$, $p_e^w$ is the number of sentences that contain n-grams $w$ related to $e$, ${\left \| p_e \right \|}_1$  is the number of n-grams in $e$, and $b$ is a smoothing parameter. We select the top $n\%$ n-gram vectors by scoring n-grams using Formula~\ref{e3}.

\subsubsection{Filter Initialization :} First of all, we embed each n-gram into a vector representation using BERT. Since filters in CNNs are insufficient on the amount to account all selected n-grams, we only extract the cluster vectors to represent the generalized cause/effect features. We perform k-means clustering to obtain clusters. Here, the cluster vector is defined as the centroid vector of the cluster. Considering that non-causal n-grams exist in the sentence, we purposely leave some blank. We do not fully fill the convolution filters with centroid vectors and randomly initialize the rest weights of the filters.

%To be continued
\begin{equation}
c_i^n = f(W\cdot h_{i:i+n-1}+b)
\end{equation}
Namely, $W$ is the weight of initialized filter, $b$ is a bias term and $f$ is a nonlinear function such as
sigmoid, ReLU. As a single cluster is often not enough to describe the overall information, we need to cluster with different length scales. To capture the features of causal events at different scales, we use parallel convolution operation with varying windows. For example, convolution window sizes of  $n_0,n_1,n_2$ are chosen when there are three convolution layers. For example, when length equals to 4, the model tends to extract \begin{CJK*}{UTF8}{gbsn}``股价下跌''\end{CJK*} (share prices fell) and \begin{CJK*}{UTF8}{gbsn}``销量减少''\end{CJK*} (sales reduced) in the Financial dataset. The convolution output can be computed as the concatenation of sub-spaces:
\begin{equation}
c_i =  [c_i^{n_0} \oplus c_i^{n_1} \oplus c_i^{n_2}]
\end{equation}
\subsection{Query-Key Attention}
Query-key attention is a special attention mechanism according to a sequence to compute its representation. It has been successfully applied in many NLP tasks, such as machine translation and language understanding. In our model, the query-key attention mechanism is employed to mine inter associations between cause and effect events. After the multi-scaled convolution operation, we obtain the feature mapping for each token. Instead of max-pooling, we use the multi-head attention mechanism to check whether exist a cause-effect relation. Following Vaswani et al. \cite{vaswani2017attention}, we split the encoded representations of the sequence into homogeneous sub-vectors, called heads. The input consists of queries $Q \in \mathbb{R}^{t\times d}$, keys $K\in \mathbb{R}^{t\times d}$ and values $V\in \mathbb{R}^{t\times d}$, while $d$ is the dimension size. The mathematical formulation is shown below:
\begin{equation}
Attention\left(Q, K, V\right) = softmax\left(\frac{QK^T}{\sqrt{d}} \right)V,
\end{equation}
\begin{equation}
H_i = Attention\left(QW_i^Q,KW_i^K,KW_i^V \right),
\end{equation}
\begin{equation}
H_{\mbox{head}} = \left([H_1\oplus H_2 \oplus \dots \oplus  H_h]\right)W,
\end{equation}
where $h$ is the number of heads. Query, key, and value matrices dimension is $d/h$. We perform the attention in parallel and concatenate the output values of $h$ heads. The parameter matrices of $i$-$th$ linear projections $W_i^Q$ $\in \mathbb{R}^{n\times {(\frac{d}{h})}}$, $W_i^K$ $\in \mathbb{R}^{n\times {(\frac{d}{h})}}$, $W_i^V$ $\in \mathbb{R}^{n\times {(\frac{d}{h})}}$. In addition, the outputs of CNN and attention structure are cascaded to output token-wise contextual representations.

\subsection{BiLSTM+CRF}
Long Short Term Memory (LSTM) is a particular Recurrent Neural Networks (RNN) that overcomes the vanishing and exploding gradient problems of traditional RNN models. Through the specially designed gate structure of LSTM, the model can optionally keep context information. Considering that the text field has obvious temporal characteristics, we use BiLSTM to model the temporal dependencies of tokens. Conditional random field (CRF) can obtain tags in the global optimal chain given input sequence, taking the correlation tags between
neighbour tokens into consideration. Therefore, for the sentence $S = \left\{s_1,s_2,s_3,...,s_n\right\}$ along with a path of tags sequence
$y = \left\{y_1,y_2,y_3,...,y_n\right\}$, CRF scores the outputs using the following formula:
\begin{equation}
score(S, y) = \sum_{i=1}^{n+1}A_{y_{i-1},y_i} + \sum_{i=1}^{n}P_{i,y_i}
\end{equation}
\subsection{Training Objective and Loss Function}
The goal of our training is to minimize the loss function. The mathematical formulation is defined as below:
\begin{equation}
E = {log}\sum_{y\in{Y}}{exp}^{s(y)}-{score(s, y)},
\end{equation}
where $Y$ is the set of all possible tagging sequences for an input sentence.

\section{Experiment}
\subsection{Datasets}
We conduct our experiments on three datasets. The first dataset is the Chinese Emergency Corpus (CEC). CEC is an event ontology corpus \footnote{\url{https://github.com/shijiebei2009/CEC-Corpus}} publicly available. It consists of six event categories: outbreak, earthquake, fire, traffic accident, terrorist attack, and food poisoning. We extract 1,026 sentences mention event and causality from this dataset. Since there are few publicly available datasets, to make the fair comparison, we also conduct experiments on two in-house datasets, one called ``Financial'', which is built based on Chinese Web Encyclopedias, such as Jinrongjie\footnote{\url{http://www.jrj.com.cn/}} and Hexun\footnote{\url{http://www.hexun.com/}}. This dataset contains a large number of financial articles, including cause-effect mentions. The Financial dataset is divided into a training set (1,900 instances), validation set (200 instances), and test set (170 instances). Because few English datasets are publicly available for event causality extraction, we re-annotated the SemEval-2010 task 8 dataset \cite{hendrickx2010semeval} and evaluate our model on this dataset. Finally, we obtain 1,003 causality instances from this dataset.

\setlength{\tabcolsep}{1.0mm}
\begin{table}
\centering
\caption{Statistic details of datasets, including training, development and test sets.}\label{tab1}
\begin{tabular}{|l|c|c|c|}
\hline
\textbf{Statistics}  &  \textbf{CEC} & \textbf{Fiancial} & \textbf{SemEval2010}\\ \hline
Average sentence length &  31.14 & 57.94 & 18.54\\
Mean distance between causal events &  10.24 & 13.49 & 5.33\\
Mode of cause event length and proportion &  2(41\%) & 4(31\%) & 1(85\%)\\
Mode of effect event length and proportion &  4(27\%) & 4(23\%) & 1(91\%)\\
Average value of cause event length &  4.03 & 6.06 & 0.96\\
Average value of effect event length  &  5.41 & 6.46 & 0.98\\
\hline
\end{tabular}
\end{table}

\subsection{Experimental Settings}
We use the pre-trained uncased BERT model. We set the model hyper-parameters according to \cite{devlin2018bert}. For all datasets, we set the maximum length of sentences to $100$, the size of the batch to $8$, the learning rate for Adam to $1 \times10^{-5}$, the number of training epochs to $100$. To prevent over-fitting, we set the dropout rate to $0.5$. It is worth mentioning that the lengths for n-grams are assigned through our preliminary statistics. For n-gram clustering, we set the number of clusters equal to the dimension of convolution filters; both are 100.
\setlength{\tabcolsep}{3.2mm}
\begin{table}[t]
\caption{Average F1-scores (\%) of joint event causality extraction using different models. Boldface indicates scores better than the baseline system. The lengths of high-frequency cause/effect events are 2, 4, and 1 for the CEC, Financial and SemEval2010 dataset, respectively. }
\label{table1}
\centering
\begin{tabular}{|ll|ccc|}
\hline
\textbf{Model}            &                & \textbf{CEC} & \textbf{Financial} & \textbf{SemEval2010}\\ \hline
IDCNN+CRF & \cite{strubell2017fast}                & 68.26       & 71.81     & 68.59  \\
BiLSTM+CRF & \cite{huang2015bidirectional}        & 68.74  & 74.75    & 73.20   \\
CNN+BiLSTM+CRF & \cite{ma2016end}              & 71.68      & 74.31     & 74.20 \\
CSNN  &\cite{jin2020inter}                            & 70.61
     & 74.59     & 73.71     \\ \hline
BERT+CSNN (baseline) & & 74.61       & 76.23       & 75.69       \\ \hline
CISAN               &   & 72.49      & 75.99     & 74.20      \\
BERT+CISAN (unigram)  &&\bf{75.26} & 76.07  &\bf{77.65} \\
BERT+CISAN (bigram) &  & \bf{75.93} & \bf{76.70}  &\bf{77.26} \\
BERT+CISAN (trigram)  &&74.45 & \bf{76.29}  &\bf{77.14} \\
BERT+CISAN (quagram)  &&\bf{75.27} & \bf{77.09} &\bf{77.35}  \\ \hline
\end{tabular}
\end{table}

\subsection{Results and Analysis}
We make a comparison between our model and previous models and also conduct additional ablation experiments to prove the effectiveness of our model. Table~\ref{table1} reports the results on CEC, Financial and SemEval2010 datasets. Following previous works, we use the standard F1-score as the evaluation metrics.

\textbf{IDCNN+CRF} \cite{strubell2017fast}: A modified CNN model, using Iterated Dilated Convolutional Neural Networks, which permits fixed-depth convolutions to run in parallel across documents to speedup while retaining accuracy comparable to the BiLSTM-CRF.

\textbf{BiLSTM+CRF}: This is a classic sequential labeling model \cite{huang2015bidirectional}, which models context information using Bi-LSTM and learns sequential tags using CRF.

\textbf{CNN+BiLSTM+CRF}: This model \cite{chen2017improving} uses CNN to enhance BiLSTM+CRF to capture local n-gram dependencies.

\textbf{CSNN}: The variation  \cite{jin2020inter} modifies the basic architecture of BiLSTM+CRF with additional CNN layer to address n-gram inside features, and the self-attention mechanism to address n-gram outside features.

\textbf{CISAN}: Our model with convolutional semantic-infused filters and key-query attention, using GloVe word embeddings.

\textbf{BERT+CSNN}: BERT serves as the encoder for CSNN, replacing the GloVe embedding layer.

\textbf{BERT+CISAN}: Our model utilizes convolutional semantic-infused filters, key-query attention incorporating with the BERT encoder.
\subsection{Effectiveness of semantic infusion}
To verify that the effectiveness of BERT and convolution semantic infusion both contribute to extraction, we construct another strong baseline BERT+CSNN. The result on BERT+CISAN produces a decent f1 score, which is 1.32\%, 0.86\%, and 1.96\% higher than BERT+CSNN on CEC, Financial, and SemEval2010, to show the effectiveness of n-grams and knowledge produced by the pre-trained model.

As described, the length of n-grams serves as a determining hyper-parameter, which reflects the frequency information of event length in our model. To prove this hypothesis, we assign the different value of lengths to verify the effectiveness and investigate the influence on the results. Table.\ref{table1} indicates that the best length of high-frequency cause/effect n-grams for the CEC, Financial and SemEval2010 dataset are 2, 4 and 1, respectively.
\begin{CJK*}{UTF8}{gbsn} Intuitively, the causal events in Financial are very likely to be described as a phrase whose length=4 like ”股价下跌(Share prices fell)” and ”销量减少(sales reduced)”. Thus, We think that convolution is more sensitive to such events. When using our method with the length of 4 for n-grams, we get the best results compared to other lengths, which demonstrates the effectiveness of semantic infusion.
 \end{CJK*}
% By initialising the convolution kernel, we incorporate the prior knowledge into the model to better mine the semantics. Furthermore, the model achieves state-of-the-art effect on three datasets when integrated with pre-trained representation BERT.
\begin{figure}[t]
%\centering

\begin{minipage}{0.49\linewidth}
\includegraphics[width=1.0\linewidth]{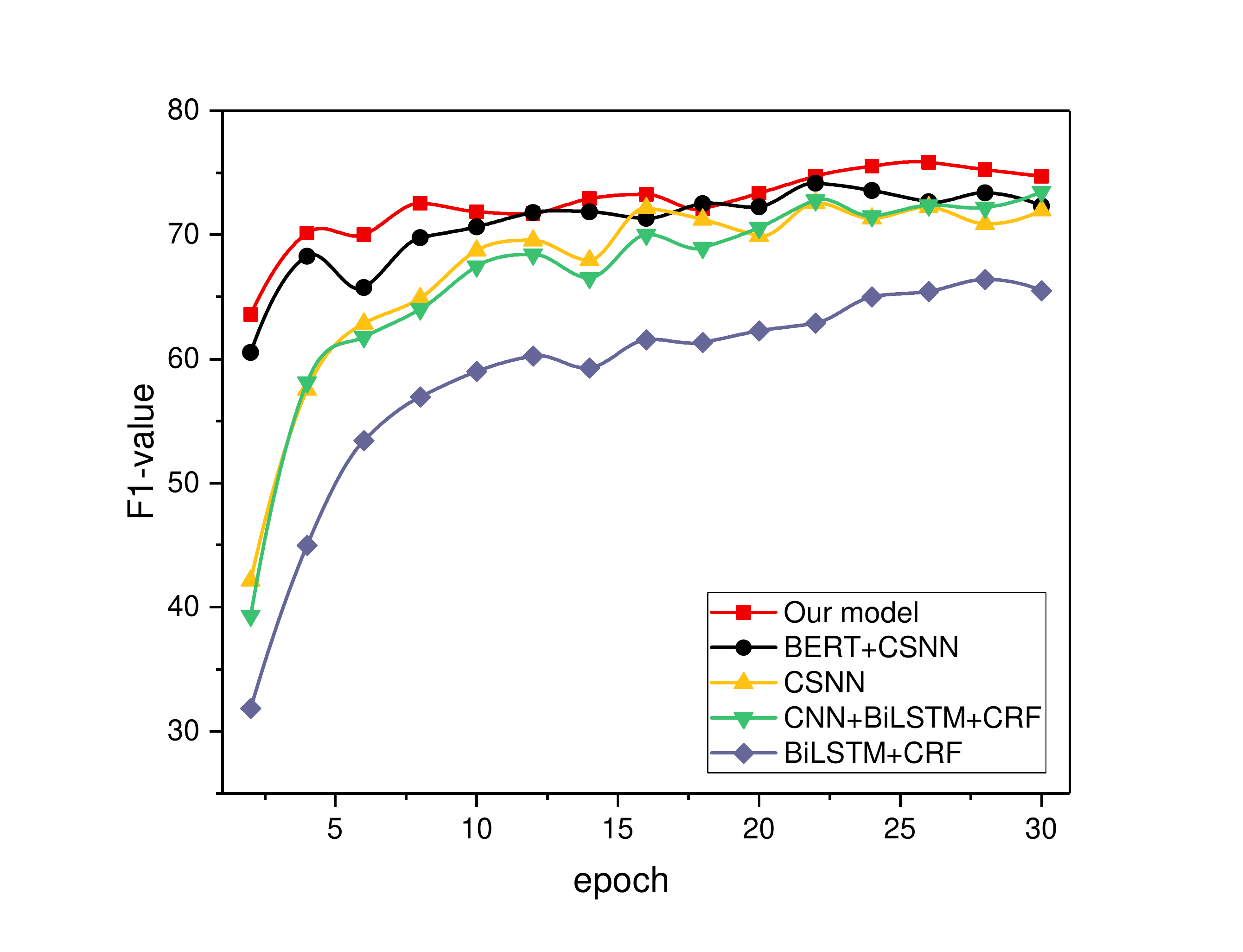}
%\centerline{(a)}
\caption{F1-score on test dataset of Financial w.r.t training epoch.}
\label{fig3}
\end{minipage}
\quad
\begin{minipage}{0.49\linewidth}
%\centering
\includegraphics[width=1.0\linewidth]{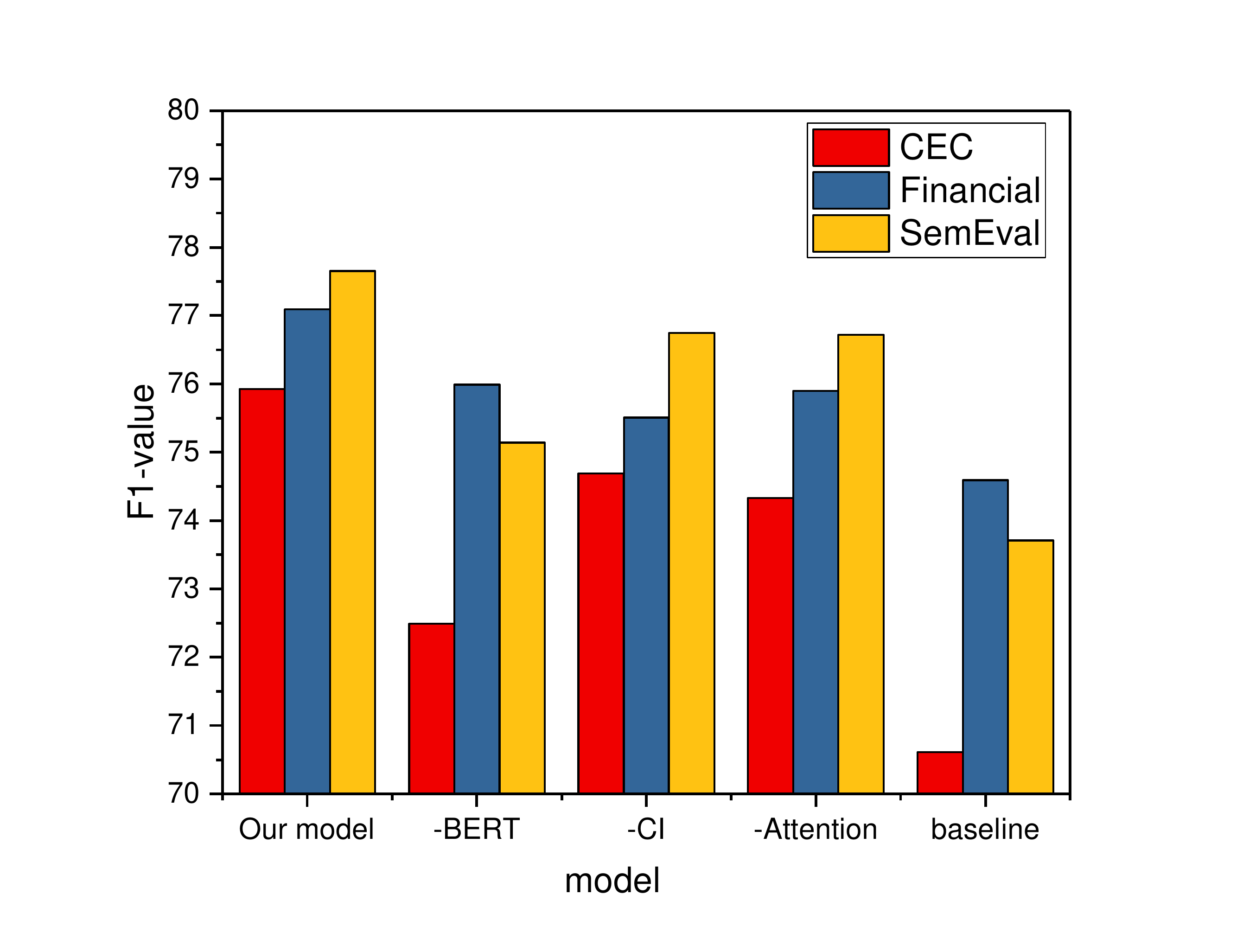}
\caption{Ablation analysis of our model in three datasets.}
%\centerline{(b)}
\label{fig4}
\end{minipage}
\end{figure}

\subsubsection{Comparison w.r.t. epoch.}
We further compare the convergence speed of models. As shown in Figure~\ref{fig3}, the NON-BERT model starts with a low F1 score from the very beginning. Reversely, our model gains a promising consequence when epoch\textless10.
Moreover, compared with the strong baseline represented by BERT, our model also obtain obvious advantages.
This suggests that prior knowledge helps the model to surpass others at few training epochs.

\subsubsection{Ablation experiments.}
We verify whether each component has a positive effect on the model by removing (BERT, CI, Attention) layer in order. The results of experiments launched on three datasets are shown in Figure~\ref{fig4}. The contribution of BERT layer is most significant for the reason that BERT has sufficient pre-training and enriched with rich semantic knowledge. Moreover, the model is promoted by extra gains of CI and attention layers. The CI layer incorporating knowledge related to cause/effect event improves the F1 score. The attention layer applied is also efficient in removing off the noise. In conclusion, all these layers simultaneously contribute to obtaining SOTA performance.

%\textbf{character vs word.} In the Chinese NLP task, there is two granularity of word-based and character-based. Generally, the character-based convolutional neural network achieved a better result than word-based one \cite{zhang2015character} in Chinese NER. Thus, we compared the performance of the baseline model based on words/character and found that the character-based model is better than the word-based one (2.86\% and 1.25\% improvement on the Financial and CEC dataset respectively). On the other hand, BERT is the character-based model in the Chinese dataset, so we chose the character-based model in the Chinese dataset.

\section{Conclusion}
This paper introduces a novel n-gram-based neural solution for joint casual relation and event extraction. The model looks up high-frequency cause/effect n-grams in the sentence by partially encoding the causal features into convolution filters. In practical, we model the associations of n-gram pairs to find the potential cause-effect pairs. As a result, our proposal achieves significant improvements compared with baseline. This work shows the feasibility of further enhancing
intra- and inter- knowledge around n-grams in guiding neural extractor. In the future, a potential direction is to adopt our method to few-shot learning of coarse-to-fine causality extraction. We are also interested in multiple cause-effect pairs extraction rather than only extracting one cause-effect event pair per time.


\bibliographystyle{splncs04}
\end{document}